\documentclass[letterpaper, 10 pt, conference]{ieeeconf}  
\usepackage{graphicx}
\usepackage{xcolor}
\usepackage{amsmath}

\IEEEoverridecommandlockouts                              

\title{\LARGE \bf
Optically Sensorized Electro-Ribbon Actuator (OS-ERA)
}

\author{Carolina Gay$^{1}$, Petr Trunin$^{1,2}$, Diana Cafiso$^{1}$, Yuejun Xu$^{3}$, Majid Taghavi$^{3}$, Lucia Beccai$^{1}$
\thanks{$^{1}$ C. Gay, P. Trunin, D. Cafiso and L. Beccai are with the Soft BioRobotics and Perception Lab of the Istituto Italiano di Tecnologia, Via Morego 30 16163, Genova, Italy (carolina.gay@iit.it, diana.cafiso@iit.it, lucia.beccai@iit.it).
}%
\thanks{$^{2}$ P. Trunin is with the Open University Affiliated Research Centre at
Istituto Italiano di Tecnologia (ARC@IIT), Istituto Italiano di Tecnologia,
16163, Genova, Italy (petr.trunin@iit.it).
}
\thanks{$^{3}$ Y. Xu and M. Taghavi are with the Department of Bioengineering, Imperial College
London, SW7 2BX London,U.K.
(yuejun.xu22@imperial.ac.uk, m.taghavi@imperial.ac.uk).
}
}

\begin{document}

\maketitle
\thispagestyle{empty}
\pagestyle{empty}
\begin{abstract}

Electro-Ribbon Actuators (ERAs) are lightweight flexural actuators that exhibit ultrahigh displacement and fast movement. However, their embedded sensing relies on capacitive sensors with limited precision, which hinders accurate control. We introduce OS-ERA, an optically sensorized ERA that yields reliable proprioceptive information, and we focus on the design and integration of a sensing solution without affecting actuation. To analyse the complex curvature of an ERA in motion, we design and embed two soft optical waveguide sensors. A classifier is trained to map the sensing signals in order to distinguish eight bending states. We validate our model on six held-out trials and compare it against signals' trajectories learned from training runs. Across all tests, the sensing output signals follow the training manifold, and the predicted sequence mirrors real performance and confirms repeatability. Despite deliberate train–test mismatches in actuation speed, the signal trajectories preserve their shape, and classification remains consistently accurate, demonstrating practical voltage- and speed-invariance. As a result, OS-ERA classifies bending states with high fidelity; it is fast and repeatable, solving a longstanding bottleneck of the ERA, enabling steps toward closed-loop control.

\end{abstract}

\section{Introduction}

\begin{figure}[!htb] \centering \includegraphics[width=1\columnwidth]{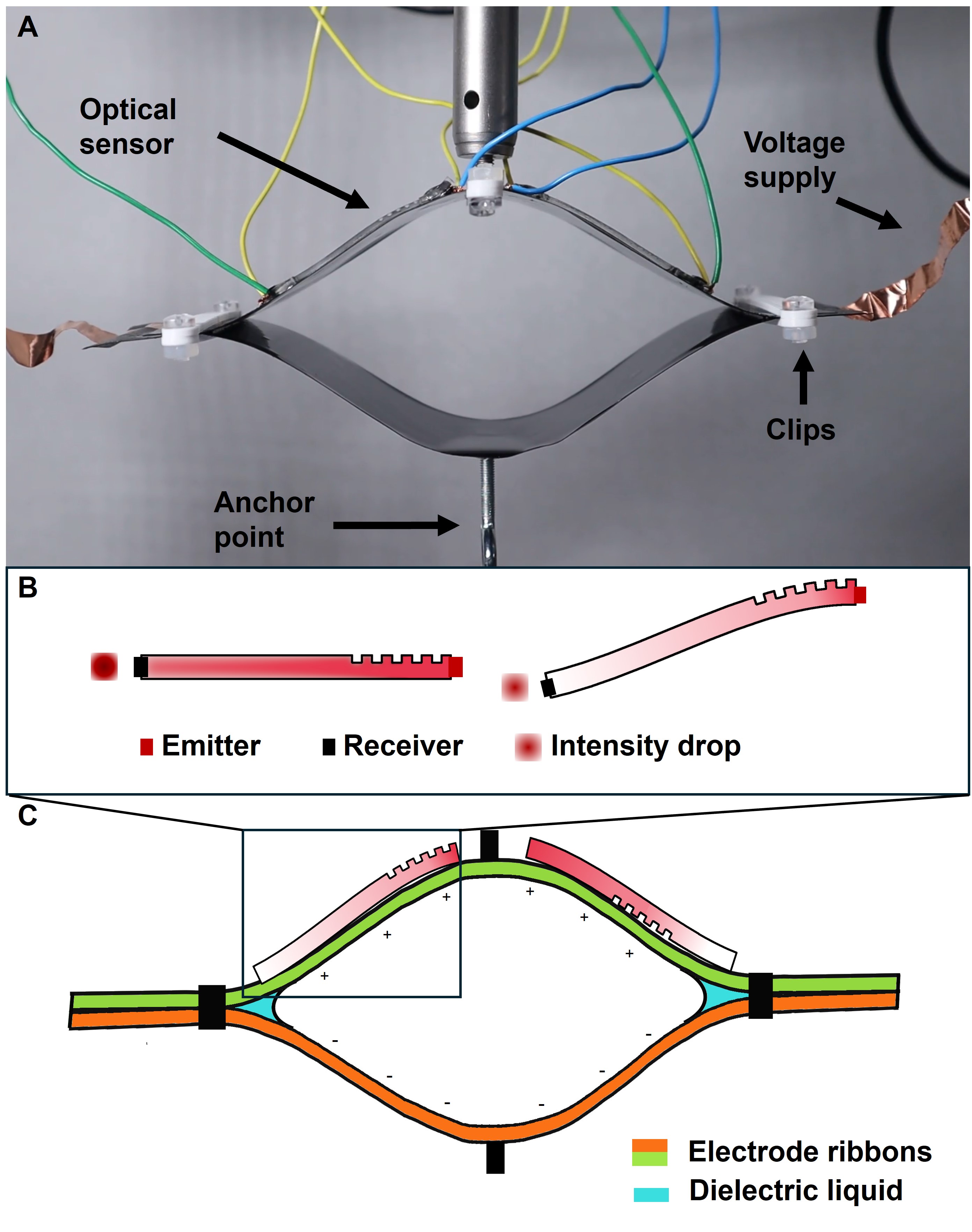} \caption{A. Optically Sensorized Electro-ribbon actuator during movement. B. Working principle of the bending optical sensor C. Working principle of ERA.} \label{fig:figure1} \end{figure}

Electro-ribbon actuators (ERA) are lightweight, flextural electrostatic actuators, capable of lifting up to 1000 times their own weight \cite{Taghavi et al 1}. Known for their exceptional contraction (close to 100\%), rapid response and human-muscle equivalent power, they are also made of inexpensive materials, representing a cost-effective and efficient solution for soft robotics \cite{Xu et al}\cite{Taghavi et al 2}. An ERA actuator consists of two thin conductive ribbons clipped together at their ends, and separated by an insulating layer (Figure~\ref{fig:figure1}-A,C). When the voltage is applied, the two electrodes get opposite charges, and an electrostatic force is generated. The dielectric liquid, applied as a drop at the clamping points, amplifies this electrostatic attraction, drawing the two ribbons closer to the zipping point. This process is called dielectrophoretic liquid zipping \cite{Taghavi et al 1}. The ERA zips rapidly only when the voltage reaches the threshold (pull-in voltage). Such soft actuators show remarkable suitability for active origami applications. Owing to their ability to be directly integrated into fold lines, they have been successfully employed in a variety of electroactive origami designs, including adaptive grippers, foldable fans, locomotion robots, and paper-based mechanisms \cite{hou2024}. One notable application of electro-ribbon actuators is their role as integrated artificial muscles due to their muscle-like characteristics \cite{xu2025}, which can be used to emulate complex biological motions, such as the wing flapping of bumblebees \cite{chong2022}.
One open issue is that ERA exhibit pull-in instability,
which, together with the nonlinear movement and the inherent compliance, makes them challenging to control and sets the need for accurate, real-time sensing. Due to this, the conventional feed-forward control scheme is not feasible, so some closed-loop strategies have been developed. For instance, a modified proportional-integral closed-loop controller (Boost-Pi \cite{Diteesawat et al}) was proposed, but its use is limited to only staircase and oscillatory control performances. Castro et al. \cite{Castro et al} developed a model (lumped-parameter model) and simulated the ERA behavior as a deformable and flexible beam, in order to predict the actuator's deformation under external loads with less than 5\% error. Another solution consists of self-sensing, exploiting the specific configuration of the ERA. In particular, Bluett et al. \cite{Bluett et al} analyzed the relationship between the capacitance and deformation of the electro-ribbon to obtain a closed-loop position control system. Nevertheless, this method requires the load to be known in advance as it is part of the model. Furthermore, a mechanical model for predicting the quasi-static behaviour of ERA has also been developed \cite{Xu et al}.
Unlike previous solutions, which rely on predictive motion models to estimate the actuator’s state based on its input signals or kinematic assumptions, the use of external sensors can provide direct feedback and real-time state classification.

Among different solutions, optical sensors are suitable candidates, due to their lightweight electronics and low susceptibility to electromagnetic interference noise \cite{Shen et al.}. These properties fulfill fundamental requirements for ERA's sensorization.
Therefore, in this work, we introduce a new sensing approach for these actuators based on soft waveguides: optically sensorized electro-ribbon actuator (OS-ERA). In particular, 3D printed waveguides with a superficial pattern are employed. Their working principle is shown in (Figure~\ref{fig:figure1}-B). The emitter emits light longitudinally through the soft, transparent material. At the end of the waveguide, a receiver is placed. When the waveguide bends, light refraction increases due to scattering caused by the wells, the receiver detects a decrease in light intensity that can be read as a voltage drop \cite{trunin2025}.
In this work, the geometry of the sensors and the positioning of the wells are designed to detect the bending states of the actuator during loading. The combination of the output signals of two sensors integrated in the actuator is used to train a supervised machine learning model based on a Support Vector Machine (SVM) \cite{akdag2024}\cite{zhang2001}\cite{chandaka2009}\cite{bose2019} to classify the ERA’s motion into eight distinct states during loading.

\section{Curvature sensor for ERA}

\subsection{Curvature analysis and design}

\begin{figure}[!htb]
\centering
  \includegraphics[width=1\columnwidth]{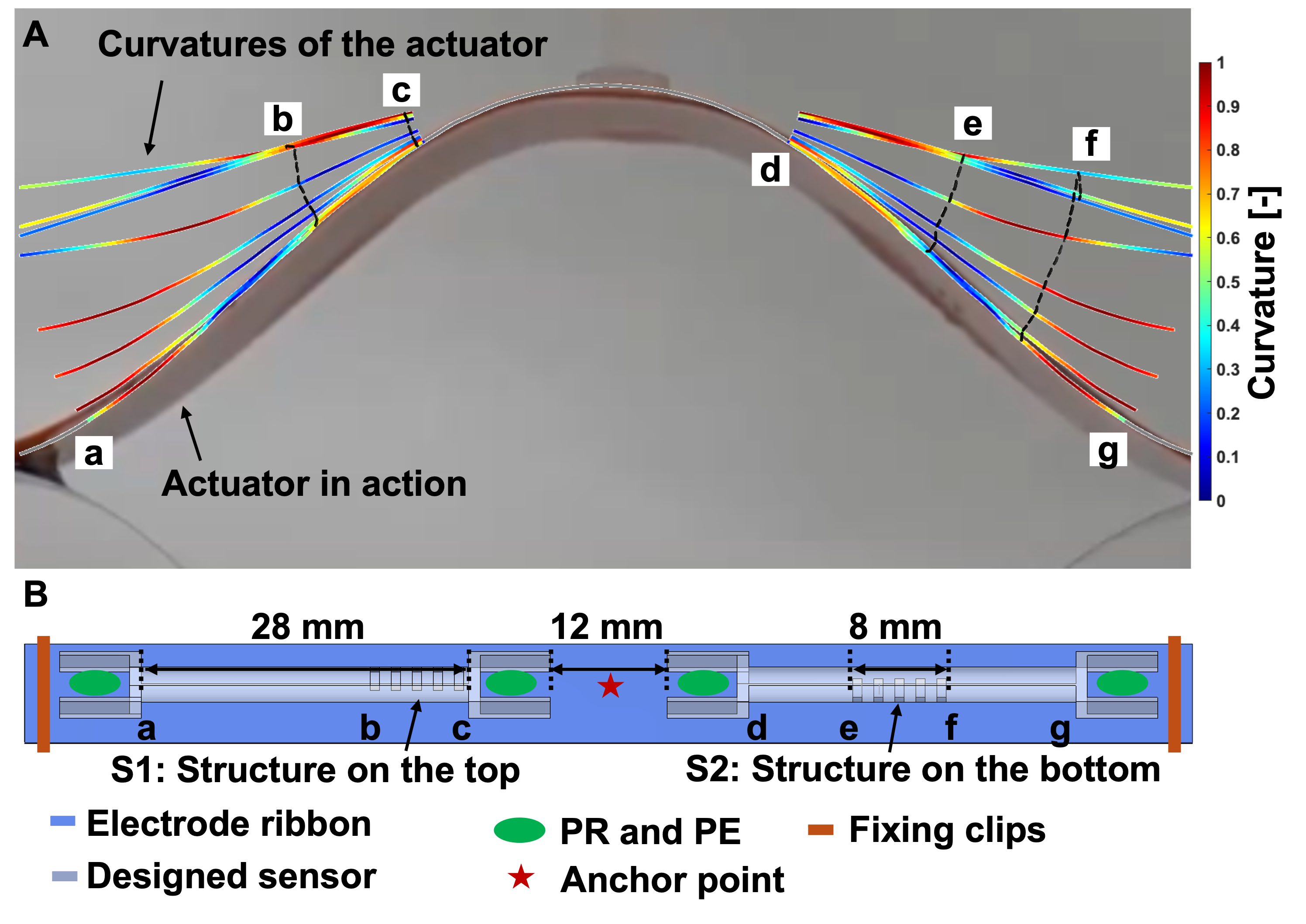}
  \caption{A. Curvature of the two areas to sensorize of the upper electro-ribbon during the loading (20 g mass for 51.75 mm with a voltage of 8 kV). B. Design of the upper ribbon surface equipped with the soft optical sensors, S1 and S2. Two emitters are positioned in the clips near the anchoring point, while two receivers are placed in the distal clips.}
  \label{fig:figure2}
\end{figure}

The ERA is composed of two electro-ribbons (length = 100 mm and width = 12.7 mm) attached at the extremes with two clips (Figure~\ref{fig:figure1}-A).
In the upper and lower centers, there is space for the anchoring and for the weight hook, respectively.
Given the structures of the actuator and the optical sensors, the best design to maintain a balanced weight is to place two sensors along the upper ribbon symmetrically (Figure~\ref{fig:figure2}-B). It is necessary to leave enough space for the electrical components and the anchoring, so the waveguides are designed with a beam length of 28 mm and rectangular caps (for electrical components) of 7 mm, as shown in Figure~\ref{fig:figure2}-B. The beam has a cross-section in the form of a circular segment with two equal parallel bases, derived from a circle of 3 mm radius and truncated by 0.5 mm at both the upper and lower sides.

3D printing enables the fabrication of a series of rectangular cavities (wells) along the beam’s surface, which increases light scattering and thus amplifies the signal change (Figure~\ref{fig:figure1}-B). Each structure is composed of a series of five rectangular wells (depth = 0.6 mm, width = 0.8 mm, spacing between them = 1 mm) along the beam’s base.
When placing the pattern, both its position along the waveguide’s main axis and the surface on which it is printed play a key role. Two considerations guide the choice:
(i) the superficial wells increase local sensitivity; therefore, the pattern should be located where maximum sensitivity is required; and
(ii) Scattering occurs mostly on the side of the waveguide that is subjected to tensile strain during bending; hence, the pattern should be placed on the surface that is stretched rather than compressed (i.e., detection of concave versus convex curvature).

Given these considerations, an experimental analysis of ERA curvature was performed. To identify suitable locations for the wells on the optical sensors, eight representative frames were taken from an ERA loading video (ERA lifting a 20 g mass for 51.75 mm with a pull-in voltage of 8 kV and performing a contraction of 99.31\% of its length) \cite{Taghavi et al 1}. Specifically, the first captured frame corresponds to the fully extended (open) configuration, whereas the last corresponds to the fully contracted (closed) configuration. Intermediate frames were selected to represent evident transitions in the zipping motion. Therefore, eight curve positions of the upper ribbon were extracted and imported into MATLAB (MathWorks, \textit{MATLAB R2024b}) to perform the analysis. The analysis consisted of identifying the ribbon’s sections where there was the most significant change in curvature during loading.  In this context, curvature described how sharply the ribbon bends \cite{Eppstein2024}, and was used as a key parameter to configure the best design. The curvature was calculated as follows \cite{Eppstein2024}:
\begin{equation}
    \kappa(i) = \frac{|x'(i)y''(i) - y'(i)x''(i)|}{\big(x'(i)^2 + y'(i)^2\big)^{3/2}}
\end{equation}

where $i$ is the pixel index corresponding to the points of the extracted curves, and $x$ and $y$ correspond to the two dimensions of the frames.
The curvature $\kappa$ was filtered using a second-order Butterworth low-pass filter, with a cutoff frequency equal to 2\% of the estimated sampling frequency.
The curvature was calculated for each curve.

The curvature analysis presented in this section is intended as a design-oriented tool rather than a precise geometric reconstruction of the actuator shape. Its purpose is to identify regions of the electro-ribbon that experience the largest curvature variations during loading, in order to inform the placement and orientation of the optical waveguide patterns. As such, the analysis emphasizes relative curvature changes across representative configurations, rather than absolute curvature values or full 3D shape characterization. 
A video-based, 2D frame extraction approach was adopted to keep the analysis simple and easily reproducible, without requiring external motion capture systems or complex 3D reconstruction pipelines. While this method does not capture out-of-plane deformations, it provides sufficient spatial resolution to identify consistent curvature trends along the actuator length, which is adequate for guiding sensor placement.

Figure~\ref{fig:figure2}-A shows the curvature values, normalized with respect to their maximum, for the regions selected for sensor placement across the eight loading states.
This analysis is essential for identifying which regions to sensorize: the structure is placed where the greatest change in curvature (Figure~\ref{fig:figure2}-A) is observed as the color transitions from red to blue (or vice versa) across curves, and the selected configurations maximize the ability to distinguish among different bending states.

Exploiting the symmetry of the system, two sensors with patterns on different sides and at different locations along the waveguide are used to enable separate detection of multiple curvature configurations.
Accordingly, two different areas are chosen for the structure of the two sensors.
In Figure~\ref{fig:figure2}, they are indicated as region $b\text{-}c$ for the first sensor and region $e\text{-}f$ for the second one, which are centered at 17.4 mm and 28.1 mm from the anchor point, respectively.
Figure~\ref{fig:figure2}-A shows the complexity of the actuator curvature, being concave and convex near the center and the ends, respectively.
Since the structure is effective when placed on the stretching side of the bent beam (Figure~\ref{fig:figure1}-C), for the first sensor the structure $b\text{-}c$ is located on the surface at the top, while for the second sensor the structure $e\text{-}f$ is located on the bottom (Figure~\ref{fig:figure2}-B).

\subsection{Fabrication and Data Acquisition}
\begin{figure}[!htb]
\centering
  \includegraphics[width=1\columnwidth]{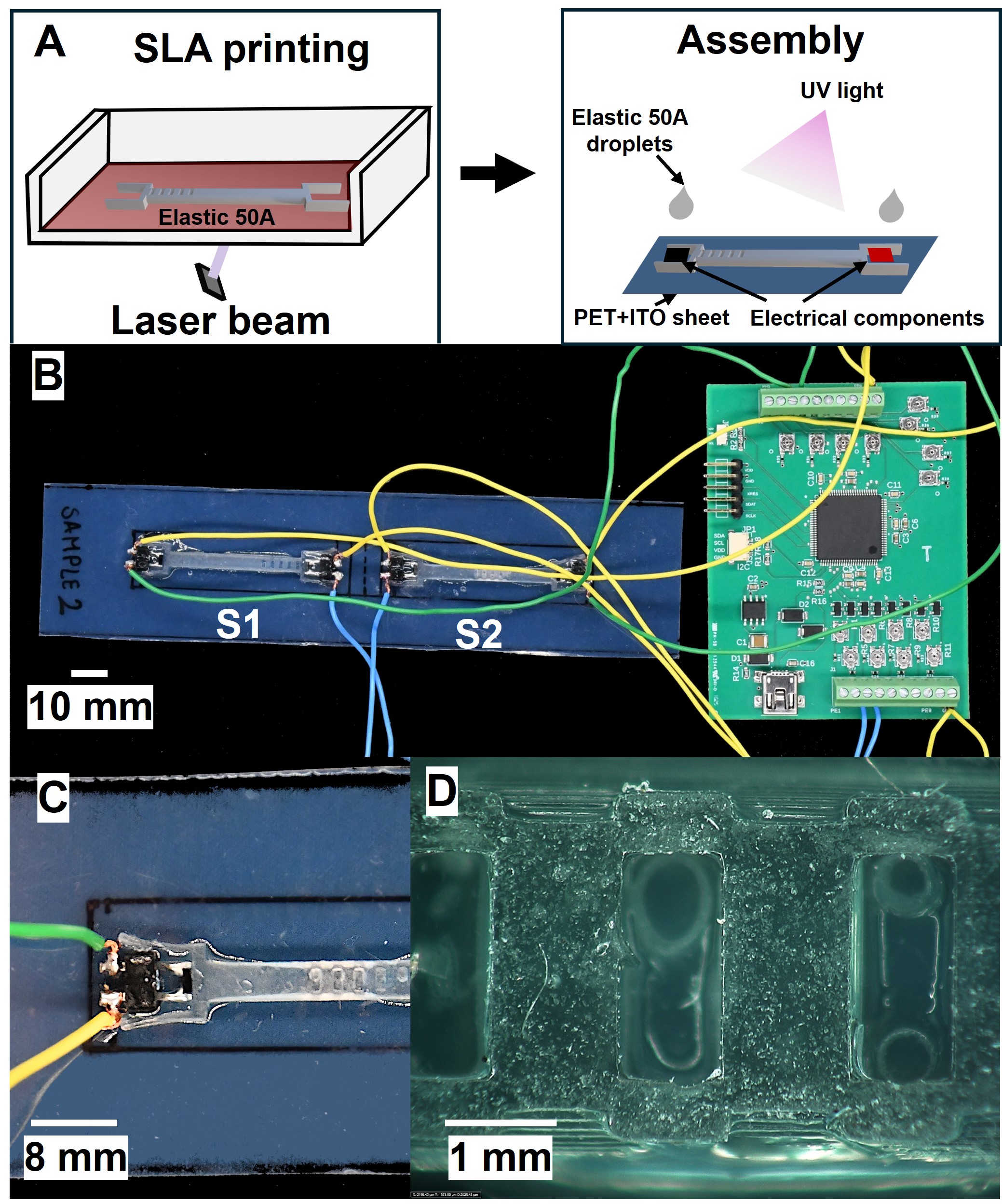}
  \caption{A. Fabrication process B. OS-ERA sample configuration C. Example of encapsulated component D. Insight image of the waveguide superficial structure}
  \label{fig:figure3}
\end{figure}

The ERA was fabricated using Indium tin oxide-coated Polyethylene terephthalate (ITO/PET, surface resistivity 60 $\Omega$/sq) with a thickness of 0.1 mm.
To fabricate one OS-ERA sample, two waveguides were printed. The waveguides were designed in Autodesk Fusion, then printed with Elastic 50A (Formlabs) and cured submerged in water under UV rays at 70 °C for 20 minutes.
Subsequently, they were attached to the PET side of the sheet (140 × 32.7 mm).
The components were successfully encapsulated with a few drops of Elastic 50A resin, as shown in Figure~\ref{fig:figure3}-A,C.
The final prototype consists of two waveguides with rectangular wells (size 0.8 × 0.6 mm).
Microscopic images (Figure~\ref{fig:figure3}-D) demonstrate the high fidelity of the printing in reproducing this superficial pattern.
In this configuration (Figure~\ref{fig:figure3}-B), the emitters are placed at the center of the sheet, while the receivers are positioned at its edges, and all these components are connected to a custom PCB.
To minimize the influence of external illumination, the board is programmed to operate with sequential activation of the emitters.
When an emitter is deactivated, the receiver measures the ambient light level, which is subsequently subtracted from the signal acquired when the emitter is active.
In addition to sequential activation, the board enables adjustment of the pull-up resistance for both emitters and receivers.
This feature allows fine-tuning of the emitted light intensity and the receiver sensitivity, thereby optimizing the signal quality and ensuring stable and reliable measurements.
For data acquisition, the PCB was connected to a serial port, and the data were read and saved using Python 3.9.

The ITO/PET substrate equipped with the optical sensors described above was used as the top beam of the electro-ribbon actuator. The bottom beam consisted of an ITO/PET sheet having same dimensions as the top substrate, but without integrated sensors. Both ITO/PET beams had their ITO-coated surfaces covered with a layer of PVC tape (Advanced Tape AT7), serving as electrical insulation. The insulating sides of the top and bottom beams were positioned facing each other and clamped together using 3D-printed PLA clips. The actuator was driven by an UltraVolt HVA series high-voltage amplifier connected through an NI DAQ. To ensure that the applied high voltage did not interfere with the sensor, the top beam was always maintained at 0 V.
At the beginning of each test, a drop of 50 cSt silicone oil (Sigma Aldrich) was applied at both zipping corners of the actuator. Simultaneously, the actuator’s deformation was recorded with a Canon EOS M50 Mark II camera for comparison with the sensor data.

\subsection{Calibration}

\begin{figure}[!htb]
\centering
  \includegraphics[width=1\columnwidth]{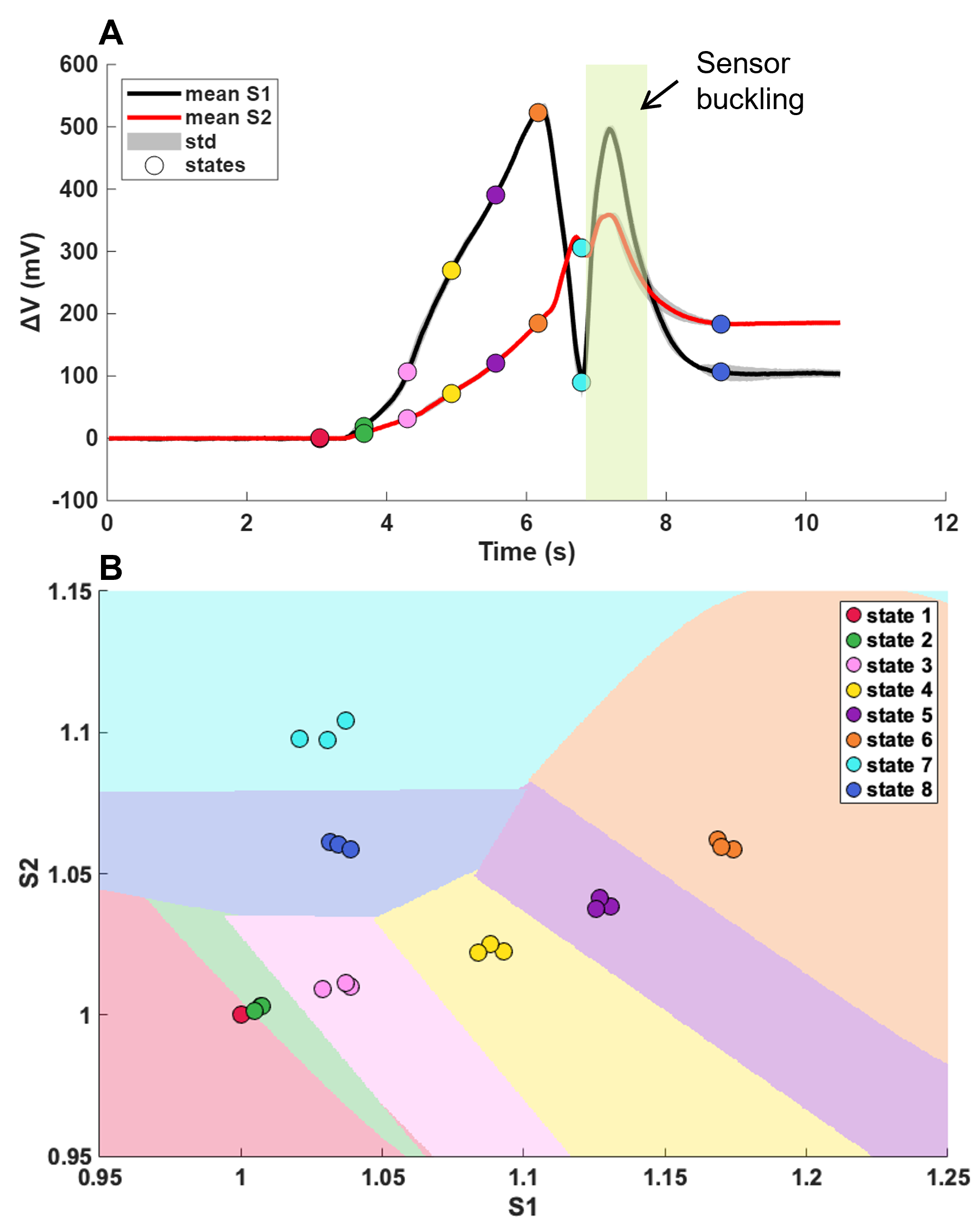}
  \caption{A. Mean and standard deviation of the three training trials (12.3 g, 4 kV) for the two sensors (in black and in red). The colored dots are equally spaced in time (except for the last one) and represent the eight selected states during actuator contraction. The green band indicates the time interval during which the sensor buckles. B. Output of the SVM model showing the two sensor signals on the axes, the colored areas and decision boundaries for the eight classes, and the corresponding training samples (dots).}
  \label{fig:figure4}
\end{figure}

The OS-ERA was tested at different conditions: 12.3 g load, 3 kV; 12.3 g load, 4 kV; 12.3 g load, 5 kV. Three trials were performed for each condition. To choose the states to classify, after removing the offset, two signals of the two optical sensors were averaged across three trials (12.3 g, 4 kV) and eight states of signals were chosen across time as shown in Figure~\ref{fig:figure4}-A. These states range from when the OS-ERA is completely open (red dot) to complete contraction (blue dot). A time interval (shown in green in Figure~\ref{fig:figure4}-A) was excluded from the analysis, as the second peak in the signals results from sensors' buckling. To build a classifier, an SVM model with a radial basis function kernel (RBF) was chosen. The RBF is particularly effective in capturing non-linear relationships between features and class labels. This kernel applies an exponential function to the squared Euclidean distance between two feature vectors \cite{amari1999}, as expressed in Equation:
\begin{equation}
K(x_i, x_j) = e^{-\gamma \|x_i - x_j\|^2} 
\end{equation}

Here, $x_i$ and $x_j$ represent the feature vectors of two data points, while $\gamma$ controls the spread of the kernel function. The classifier was implemented in Python using scikit-learn. Input features were standardized using a StandardScaler and the $\gamma$ parameter was set to $1/n_{feature}$.

 The model was trained with the three trials in which an OS-ERA sample was lifting 12.3 g with an applied voltage of 4 kV. As input to the model, the signal values were normalized with respect to the first state value for both sensors (training data size: 3 trials x 2 sensors x 8 classes). The decision boundaries found by the SVM model are represented in Figure~\ref{fig:figure4}-B in which the two axes represent the normalized signals of the two sensors, and the eight colored areas are the eight states of the actuator. The three dots per area represent the values of the three training trials for each state (whose mean across trials correspond exactly to the eight dots shown in Figure~\ref{fig:figure4}-A). During loading, the trajectory of the combined signals moves around this plane, and whenever it falls within a certain area, the model classifies that moment as the corresponding state.

\section{Results}

\begin{figure}[!htb]
\centering
  \includegraphics[width=1\columnwidth]{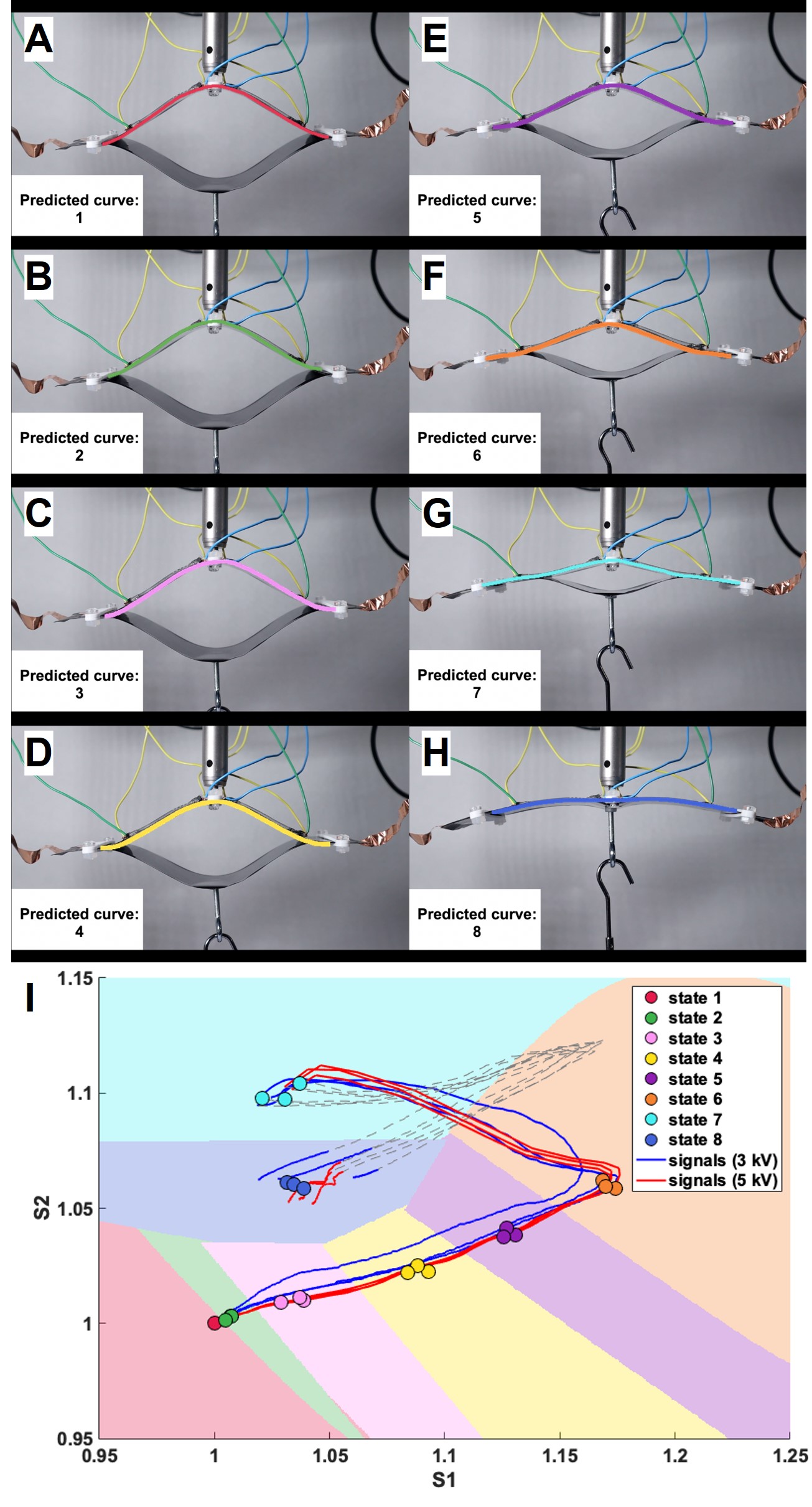}
  \caption{A-H. Prediction example during OS-ERA movement (12.3 g, 5 kV). The bottom-left number indicates the predicted state, while the colored ribbon above the actuator represents the corresponding curve (saved from training video). The images show frames extracted from the video (12.3 g, 5 kV) at the moments of each prediction. I. Signals' trajectories of all testing trials on the SVM model map. The three blue trajectories correspond to the testing trials performed at 3 kV, while the three red trajectories correspond to those performed at 5 kV. The colored dots represent the training data obtained from the three trials performed at 4 kV.}
  \label{fig:figure5}
\end{figure}

The model was validated across six trials (testing set: 6 trials × 2 sensors over 25 s) under a fixed load of 12.3 g, with three trials at 3 kV and three at 5 kV. All predictions were computed in Python 3.9 and reported as overlays of the training ERA trajectory and the testing trajectories, along with the learned decision boundaries (Video S1). In all cases, the composite signal trajectory is locked in the training manifold as shown in Figure~\ref{fig:figure5}-I, advancing along the same path with a clear voltage-dependent timing shift (3 kV slower, 5 kV faster). Figure~\ref{fig:figure5}-A-H illustrates this alignment: colored curves mark eight canonical time points from a training video, while the corresponding frames from an independent 5 kV trial fall exactly at the states the model predicted.



\section{Discussion and Conclusion}
The results obtained in this study show accurate predictions across varying loading speeds and voltages, demonstrating the robustness of the proposed sensing and learning framework. Since these sensors were specifically designed for this actuator, the reported results represent a first indication that this approach can effectively address the sensorization of ERA.

The training dataset used in this study is intentionally small, consisting of three trials acquired under a single load and voltage condition. While this naturally limits statistical generalization, the primary objective of this work is to validate the sensing design and to assess whether the resulting signal manifold remains stable under variations in actuation voltage and speed. The successful generalization to unseen voltages demonstrates that the sensing signals preserve a consistent structure, partially mitigating overfitting concerns. Expanding the dataset across a broader range of loads, geometries, and operating conditions is a necessary next step and will be the focus of future studies.

The use of a classifier was motivated by the initial objective of assessing the ability to distinguish between the open (relaxed), closed (contracted), and intermediate states. This study focuses on discrete state classification rather than continuous shape or position reconstruction. The choice of eight states was intentionally conservative and driven by the goal of demonstrating robustness and invariance rather than maximizing spatial resolution. Importantly, this limitation is not imposed by the sensing modality itself: the optical waveguides provide continuous-valued signals that can support higher-resolution classification or regression-based models.  The next step will be the development of a machine learning model capable of providing continuous state estimation, for example by using a support vector regressor.

Sensor buckling may arise from the geometry of the waveguides or from the way they are attached to the actuator. This effect can potentially be eliminated or controlled in future studies through improved design and mounting strategies. Excluding a specific time interval from the analysis might appear to reduce the robustness of the proposed approach; however, this was a deliberate choice to avoid classifying signal components that were considered noise. Instead, the analysis focused on demonstrating that the sensor signals are repeatable across different trials and voltage conditions.

One consideration is the need for a calibration procedure before operation. If the OS-ERA prototype changes, a new calibration must be performed by collecting new training data. This could become time-consuming when developing a large number of devices. Nevertheless, it would be interesting to investigate whether a single calibration model could be generalized to multiple prototypes. This would also help to evaluate how small differences in 3D-printed optical waveguides (due to manufacturing tolerances) may affect signal propagation. A valuable extension of this study could involve a simulation-based analysis aimed at correlating the geometric features of the waveguide pattern with light transmission characteristics. By systematically varying parameters such as the number of wells, their dimensions, and spatial distribution, an optimized design could be achieved \cite{trunin2025}.

Optical sensors are inherently well-suited for soft robotics; however, although the effect of external light was strongly mitigated in this work, it was not completely eliminated and could still affect the acquisition accuracy, especially for higher-resolution sensing. Future research could address this by introducing an optical coating specifically designed for the waveguides, capable of filtering ambient light without altering the mechanical or optical properties of the structure, thereby enabling its use in a wider range of applications.

Regarding the curvature analysis, the method adopted in this study is a simplified approach, relying on video frame analysis to extract and quantify the ribbon’s curvature. This simplicity aligns with the nature of the optical sensors, as the primary requirement was to identify the ribbon locations experiencing the greatest bending changes during loading. Despite its basic implementation, this method has proven sufficiently precise to guide the sensor design and it yields experimentally validated results, supporting the proposed OS-ERA concept as a practical and efficient solution for ERA sensing.

Overall, this work introduces the OS-ERA, an electro-ribbon actuator integrating soft optical waveguide sensors designed through curvature-informed placement. By combining two embedded optical sensors with a supervised learning approach, the system achieves repeatable and voltage-invariant classification of representative bending states. While the current implementation focuses on discrete state estimation rather than continuous shape reconstruction or closed-loop control, it establishes a robust proprioceptive sensing layer that addresses a longstanding sensing bottleneck in ERA-based systems. The proposed approach lays the groundwork for future extensions toward higher-resolution state estimation, multi-load generalization, and real-time closed-loop control.

\addtolength{\textheight}{-12cm}   




\end{document}